\documentclass{bvm2022} % Do NOT change this line

\begin{document}

% Do not change anything in the preamble (anything above \begin{document}) except for the specification of the bibliography file, any additional changes will be lost

% use the \selectlanguage command to select the language in which your proceedings are written

%\selectlanguage{ngerman} % German
\selectlanguage{english} % English

% Indication of the title of your contribution
% In English, \title, \subtitle, and \titlerunning are to be capitalized except for connective words. The chapters of the individual contribution as well as the titles of the bibliography entries, however, are to be written in lower case except for the first letter, proper names and abbreviations (decapitalized).
\title{Comparison of Evaluation Metrics\\
for Landmark Detection in CMR Images}
% If you write a short paper/abstract, the title must start with "Abstract:".
% \title{Abstract: Bildverarbeitung für die Medizin 2022}

% Optional specification of subtitle
%\subtitle{Guidelines for the creation of the print-ready contributions}

% titlerunning appears in the header of every second page
% LaTeX generates this automatically from your contribution title
% However, if it is too long, the message "Title Suppressed Due to Excessive Length" appears instead.
% In this case, specify an abbreviated form of the title here
\titlerunning{Evaluation of Landmark Detection in CMR Images}

% Please indicate all authors involved
% To allow us to correctly identify the last name of each author, indicate it using the \lname{} command.
% If more than one institute is involved, list the number of the institute(s) (see below) with \inst{} after the respective author. If only one institute is involved, omit this.
% Separate all authors with a comma
\author{Sven \lname{Koehler} \inst{1,2}, Lalith \lname{Sharan} \inst{1,2}, Julian \lname{Kuhm} \inst{1}, Arman \lname{Ghanaat} \inst{1}, Jelizaveta \lname{Gordejeva} \inst{1}, Nike~K. \lname{Simon} \inst{1}, Niko~M.\lname{Grell} \inst{1}, Florian \lname{André} \inst{1}, Sandy \lname{Engelhardt} \inst{1,2}}

% Enter the authors here as you want them to appear in the header
% Name only the surnames
% Depending on the number of authors involved, follow the examples below
% \authorrunning{Meier} - one author
% \authorrunning{Meier \& Müller} - two authors
% \authorrunning{Meier, Müller \& Schulze} - three authors
% \authorrunning{Meier et al.} - more than three authors
\authorrunning{Koehler et al.}

% Specify the institutes involved
% In case of participation of more than one institute, each institute shall be preceded by an ascending number with \inst{}.
% If only one institute is involved, omit the corresponding number.
% Separate individual institutes with \\
\institute{
\inst{1} Department of Internal Medicine III, Heidelberg University Hospital, Heidelberg\\
\inst{2} DZHK (German Centre for Cardiovascular Research), partner site Heidelberg/Mannheim}

%==============================================================================
% enter the e-mail address of the corresponding author through the
% command \email 
%
\email{sven.koehler@med.uni-heidelberg.de}

%\begin{acronym}
%\acro{cmr}[CMR]{Cardiac Magnetic Resonance}
%\acro{sax}[SAX]{short-axis}
%\acro{rvip}[RVIP]{right ventricular insertion points}
%\acro{tp}[TP]{True Positives}
%\acro{fp}[FP]{False Positives}
%\acro{fn}[FN]{False Negatives}
%\acro{ppv}[PPV]{Positive Predicted Value}
%\acro{tpr}[TPR]{True Positive Rate}
%\acro{ant}[Ant]{anterior}
%\acro{inf}[Inf]{inferior}
%\acro{gt}[GT]{ground-truth}
%\acro{pred}[PRED]{prediction}
%\end{acronym}

\maketitle

\begin{abstract}
% Purpose
Cardiac Magnetic Resonance (CMR) images are widely used for cardiac diagnosis and ventricular assessment. 
Extracting specific landmarks like the right ventricular insertion points is of importance for spatial alignment and 3D modelling.
The automatic detection of such landmarks has been tackled by multiple groups using Deep Learning, but relatively little attention has been paid to the failure cases of evaluation metrics in this field.
% Method
In this work, we extended the public ACDC dataset with additional labels of the right ventricular insertion points and compare different variants of a heatmap-based landmark detection pipeline. In this comparison, we demonstrate very likely pitfalls of apparently simple detection and localisation metrics which highlights the importance of a clear detection strategy and the definition of an upper-limit for localisation based metrics.
% Results
Our preliminary results indicate that a combination of different metrics are necessary, as they yield different winners for method comparison. Additionally, they highlight the need of a comprehensive metric description and evaluation standardisation, especially for the error cases where no metrics could be computed or where no lower/upper boundary of a metric exists. Code and labels: \url{https://github.com/Cardio-AI/rvip_landmark_detection}
\end{abstract}

\section{Introduction}

It is common to acquire Cardiac Magnetic Resonance (CMR) image in a short-axis (SAX) orientation, which is clinically used for ventricular assessment. However, while the normal of the multi-slice stack points along the long axis of the heart, there is a remaining degree of freedom around this axis (e.g. right ventricle could be to the left or on the lower/upper side of the image slice, Fig. \ref{3269-fig:overview} (c, d)). 
Recent works show that deep neural network models perform better if based on standardised, in plane-rotated CMR images \cite{3269-VigneaultOmegaNet2018}.
For rotation and to facilitate automated analysis and reporting, landmark points such as the two right ventricular insertion points (RVIP), need to be identified on the images. They define the septum, span over several z-slices and are important for 3D cardiac modelling, particularly for fitting bi-ventricular meshes or to determine the standard myocardial segments \cite{3269-CerqueiraAHAPaper.2002}. 

Multiple approaches have been previously adopted to tackle the task of landmark detection in this domain. \cite{3269-ourselin_artificial_2016} and \cite{3269-alansary_evaluating_2019} followed a reinforcement learning approach to detect anatomical landmarks from CMR images. \cite{3269-bujan_automatic_2012} presented a method for localising landmarks in CMR images using a cyclic motion mask. \cite{3269-wang_left_2020} proposed a model based on deep distance metric learning. \cite{3269-xue_landmark_2021} developed a CNN based solution for robust landmark detection for different CMR image contrasts. However, these works do not focus on the different metric definitions and their respective impact on assessment of model performance.

A very recent work \cite{3269-DBLP:journals/corr/abs-2104-05642} has started a timely discussion on the effects of choosing different evaluation metrics for segmentation tasks. 
In general, independent of the task, it is important to report how \emph{NA} cases are treated or whether an upper/lower bound to a metric exist or need to be defined. 

In this paper, we aim to show the influence of the definition of different apparently simple detection and localisation metrics in the particular use-case of RVIP landmark detection. We show that, depending on the metric, improved model variant comparison tailored to the post-processing task are possible. The results emphasise that more attention should be paid to evaluation metric definition and that authors are encouraged to provide more comprehensive descriptions.

\section{Materials and methods}

\subsection{Data pre-processing and ground truth}
To ensure good reproducibility and comparability of the presented approaches, we make use of the publicly available ACDC dataset \cite{3269-Bernard2018a}. This dataset consists of SAX CMR images from 100 patients and covers adults with normal anatomy and pathological cases. We manually labeled the anterior and inferior RVIPs as circular regions of 5 pixels, which we make openly available on our GitHub page.
The images are pre-processed while training as follows: 
In-plane resampling to a uniform spacing of $1.2\ts \text{mm}^2$, 
linear interpolation for images, nearest neighbour for the masks. 
Centre crop/pad to network input size of $224^2$ pixels.
The CMR images are further clipped by the $0.999$ quantile and min/max-normalised. 
The following augmentations are applied online with a probability of $80\%$: 
random grid distortion, $90^{\circ}$ rotation, shifts 
and downsampling. 
In a separate experiment (Var.$2$), we apply random histogram matching to deal with the different pixel value distributions.

\subsection{Model pipeline}
To handle slices without RVIPs we formulate this problem as a slice-based segmentation task, using a 
combined Binary Cross-Entropy and Dice loss. The network has a U-Net architecture, consisting of four down-/up-sampling blocks \cite{3269-Ronneberger.2015}.
All parameters are configurable, the implementation details and model definition are publicly available on GitHub. 
The training-subset was shuffled and split in a $4$-fold cross-validation manner with respect to the pathologies as done by \cite{3269-KoehlerSPIE2020}. We present four different model variants: the baseline model (Base), the model with histogram matching performed (Var.$1$), and two models with Gaussian distribution applied to the masks, with values $\sigma=2$ (Var.$2$), and $\sigma=4$ (Var.$3$). Each model was trained for $500$ epochs with early stopping.

As post-processing we apply a threshold $t=0.5$ on the predicted heatmaps and  retain only the biggest connected component per channel. Finally, we inverted  all operations to compute the metrics in the original CMR image space.

\subsection{Evaluation}

\begin{figure}[b]
\centering
\includegraphics[width=1\textwidth]{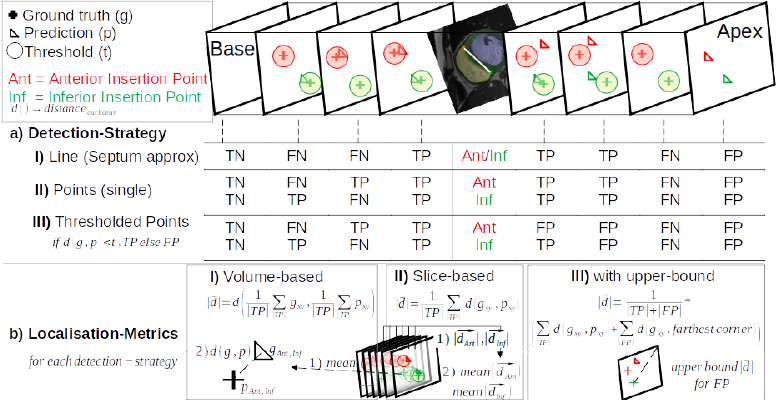}
\caption{Example cases for our three different detection strategies (a), they influence the total number of true positives, and could be combined with each localisation metrics (b).}
\label{3269-fig:evaluation}
\end{figure}

\emph{Detection-based metrics} The crucial aspect in detection based metrics is \emph{how} the True Positives (TP), False Positives (FP), and False Negatives (FN) are actually defined.
This definition may be heavily dependent on the use-case.
The Positive Predictive Value (PPV) $PPV = TP/(TP + FP)$ and the True Positive Rate (TPR) $TPR = TP/(TP + FN)$ can then be derived. We analyse three different evaluation strategies to compute the TP, FP, and FN. (1) Firstly, we adopt a line-based strategy, where we only consider the slices where both RVIPs are detected (Fig. \ref{3269-fig:evaluation} (a)(i)). This case is relevant for rotational alignment of the images, as the axis between the anterior (Ant) and inferior (Inf) RVIP is only defined when both the landmarks are predicted. In this case, only slices that contain both points in the ground-truth (GT) and predicted (PRED) are considered as TP. When only one of the points is predicted, it is considered as a FN. The slices without GT points but two predicted points are considered as FP. (2) In the second case, we adopt a point-based detection strategy, where single predicted points are also considered in the computation, i.e. the evaluation is done for each individual landmark ($TPR_{Ant}$, $TPR_{Inf}$, similarly $PPV_{Ant}$, $PPV_{Inf}$) (Fig. \ref{3269-fig:evaluation} (a)(ii)). (3) Fig. \ref{3269-fig:evaluation} (a)(iii) shows the third variant, a thresholded point-based strategy. Here, we define a radius of $15mm$ around the GT points as done in \cite{3269-wang_left_2020}. Points inside this radius are counted as TPs, and all others are FPs.

\begin{figure*}[b]
\centering
  \includegraphics[width=0.9\textwidth]{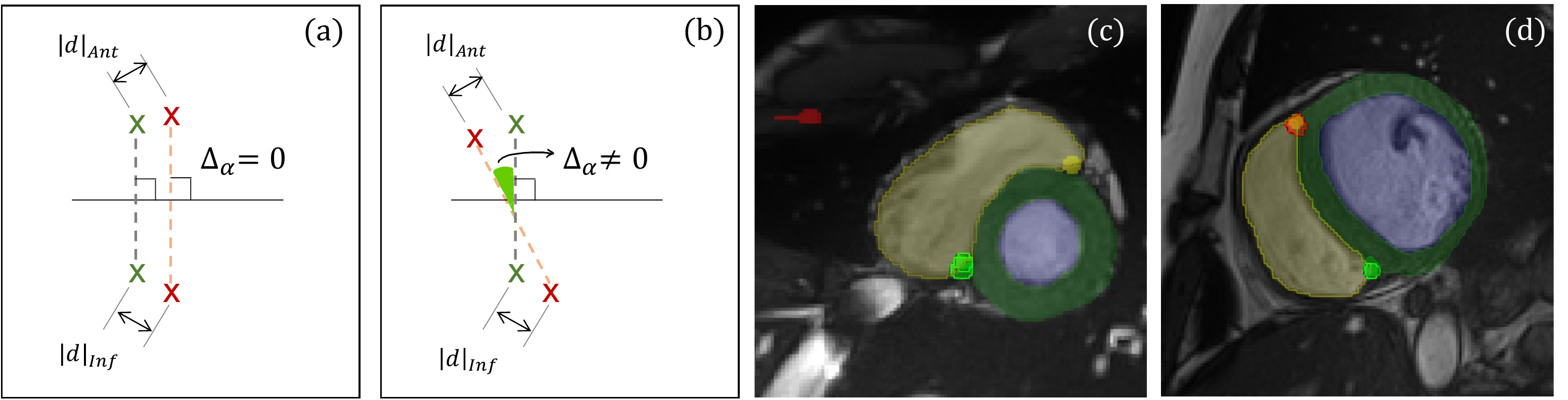}
\caption{(a) An example where the mean distance and the mean angle provide different comparisons. (b) Distance metrics remain the same whereas angle differs compared to (a). (c) Shows an example failure case prediction for the Ant RVIP on a rotated CMR and (d) a good prediction.}
\label{3269-fig:overview}
\end{figure*}

\begin{table}
\centering
\caption{Comparison of localisation metrics for different experiments (Base: Baseline, Var.$1$: + Hist. matching, Var.$2$: + Gauss $\sigma=2$, Var.$3$: + Gauss $\sigma=4$).}
\begin{tabular*}{\textwidth}{l@{\extracolsep\fill}lcccccc}
\hline
Detection-Strategy& & \multicolumn{3}{c}{(i) Line} & \multicolumn{2}{c}{(ii) Points}\\
& Exp. & $|d|_{Ant}$ $\downarrow$ & $|d|_{Inf}$ $\downarrow$&
$\Delta_{\alpha}$ $\downarrow$&
$|d|_{Ant}$ $\downarrow$& $|d|_{Inf}$ $\downarrow$\\
\hline
    \multirow{4}{1.5cm}{(i)\\Volume-\\based}
    & Base & $5.92\pm{4.83}$ & $3.86\pm{5.32}$ & $3.80\pm{4.09}$& $7.16\pm{6.88}$ & $5.79\pm{7.17}$\\
    & Var.1 & $\emph{5.58}\pm{6.25}$ & $4.16\pm{5.75}$ &$4.60\pm{7.12}$& $6.88\pm{7.71}$ & $4.86\pm{6.51}$\\
    & Var.2 & $6.26\pm{7.08}$ & $3.54\pm{3.83}$ &$4.13\pm{5.24}$& $6.93\pm{8.06}$ & $5.40\pm{9.08}$\\
    & Var.3 & $5.86\pm{4.95}$ & $\emph{3.33}\pm{3.47}$ &$\emph{3.67}\pm{3.41}$& $\emph{6.67}\pm{5.72}$ & $\emph{4.17}\pm{5.57}$\\

    \multirow{4}{1.5cm}{(ii)\\Slice-\\based}
    & Base & $4.42\pm{5.66}$ & $3.96\pm{7.07}$ &$\emph{2.70}\pm{3.09}$& $5.08\pm{9.04}$ & $3.89\pm{6.96}$\\
    & Var.1 & $\emph{3.79}\pm{7.20}$ & $3.02\pm{4.39}$ &$3.31\pm{5.77}$& $\emph{4.05}\pm{7.67}$ & $3.00\pm{4.08}$\\
    & Var.2 & $3.88\pm{4.97}$ & $3.12\pm{7.10}$ &$3.63\pm{7.51}$& $4.08\pm{5.47}$ & $3.51\pm{8.14}$\\
    & Var.3 & $4.42\pm{5.67}$ & $\emph{2.48}\pm{2.20}$ &$3.81\pm{9.00}$& $4.58\pm{6.77}$ & $\emph{2.71}\pm{2.89}$\\

    \multirow{4}{1.5cm}{(iii)\\Slice-\\based, \\$\uparrow$-bound}
    & Base & $50.33\pm{65.01}$ & $49.68\pm{65.98}$ & $30.85\pm{39.20}$& $35.05\pm{46.46}$ & $29.93\pm{59.14}$\\
    & Var.1 & $\emph{37.07}\pm{46.70}$ & $\emph{36.83}\pm{45.74}$ & $\emph{24.25}\pm{29.92}$& $\emph{29.62}\pm{42.46}$ & $\emph{14.80}\pm{28.56}$\\
    & Var.2 & $48.53\pm{64.66}$ & $47.58\pm{63.89}$ &$30.79\pm{38.75}$& $36.82\pm{57.30}$ & $27.39\pm{46.82}$\\
    & Var.3 & $55.05\pm{74.88}$ & $53.76\pm{75.66}$ &$34.66\pm{44.92}$& $39.44\pm{61.48}$ & $34.57\pm{58.76}$\\
\hline
\end{tabular*}
\label{3269-tab:distances}
\end{table}

\textit{Localisation-based metrics} Computing the Euclidean distance ($|d|$) between the GT and PRED point is a common localisation metric. As for this metric, no upper bound exists, but slices without a detected point must be handled. We analyse different variants to compute the distance and penalise false predictions. (1) In the first method, we compute a volume-based distance, where the mean location of the GT and PRED points are computed across all TP slices. Then, the distance between the mean points is computed (Fig. \ref{3269-fig:evaluation} (b)(i)). (2) In the second method, the distances are computed in a slice-based manner, considering also single detected points, following we take the mean of the distances per volume (Fig. \ref{3269-fig:evaluation} (b)(ii)).

Computing the distances with a volume-based method reduces the outlier bias and yields a good approximation of the septum orientation.
However, in these two methods the slices without a landmark prediction are not accounted for in the distance computation and therefore not penalised. To account for this, in our third method, (3) we apply an upper bound ($\uparrow$ bound) to the cases without a prediction (FN). The upper bound distance is computed as the distance from the GT points to the farthest corner of the image (Fig. \ref{3269-fig:evaluation} (b)(iii) and Fig. \ref{3269-fig:overview} (a)). For each method we present the results ($|d|_{Ant}$ for anterior points, $|d|_{Inf}$ for inferior points) considering the slices where both the landmarks are detected and only single landmarks are detected, respectively (Tab. \ref{3269-tab:distances}). 

Finally, towards the motivation of rotational alignment of the CMR images based on the septal wall, we calculate the differences of the mean septum angle ($\Delta_{\alpha}$). We define this angle as the clock-wise angle between the $x$-axis and the septal wall (Fig. \ref{3269-fig:overview} (b)).
For the $\uparrow$ bound column we used an upper bound of 180$^\circ$, as this reflects the maximum variation noticed in this dataset.

\section{Results}
Figure \ref{3269-fig:overview} (c,d) shows one good/failure case of this landmark detection pipeline. Starting with the intuitive volume-based localisation metrics (i) in Table \ref{3269-tab:distances}, it seems that Var.$3$ performs best in nearly all metrics. If we extend this view to the slice-based localisation metrics (ii), we see that metrics improved for all experiments (up to $3\ts \text{mm}$ for $|d|_{Ant}$, $1\ts \text{mm}$ for $|d|_{Inf}$). Simultaneously, the ranking changed. There is no clear winning method, each method performs best in one metric. One problem of the volume-(i) and slice-based (ii) methods is the lack of a fair FN case penalty. This becomes evident if we repeat the slice-based evaluation but, with an $\uparrow$ bound handling for FN cases (Fig. \ref{3269-fig:evaluation} (b)(iii)). When a boundary is applied to the localisation metric, Var.$1$ with the histogram matching performs better in all metrics for both anterior and inferior points (Tab. \ref{3269-tab:distances} (iii)).

\begin{table}
\centering
\caption{Comparison of detection metrics for different experiments (Base: Baseline, Var.$1$: + Hist. matching, Var.$2$: + Gauss $\sigma=2$, Var.$3$: + Gauss $\sigma=4$).}
\begin{tabular*}{\textwidth}{l@{\extracolsep\fill}lcccccc}
\hline
Detection-Strategy& \multicolumn{2}{c}{(i) Line} & \multicolumn{4}{c}{(ii) Points}  \\
Exp. & $TPR$ $\uparrow$& $PPV$ $\uparrow$& $TPR_{Ant}$ $\uparrow$& $TPR_{Inf}$ $\uparrow$& $PPV_{Ant}$ $\uparrow$& $PPV_{Inf}$ $\uparrow$\\ \hline
Base & $0.84\pm{0.22}$ & $0.84\pm{0.22}$ & $0.89\pm{0.16}$ & $0.91\pm{0.19}$ & $0.80\pm{0.22}$ & $0.77\pm{0.24}$\\
Var.1 & $\emph{0.88}\pm{0.16}$ & $\emph{0.85}\pm{0.19}$ & $\emph{0.91}\pm{0.15}$ & $\emph{0.96}\pm{0.10}$ & $0.79\pm{0.21}$ & $0.79\pm{0.21}$\\
Var.2 & $0.85\pm{0.21}$ & $\emph{0.85}\pm{0.23}$ &	$0.88\pm{0.19}$ & $0.92 \pm{0.16}$ & $\emph{0.81}\pm{0.23}$ & $\emph{0.80}\pm{0.22}$\\
Var.3 & $0.82\pm{0.25}$ & $0.83\pm{0.26}$ & $0.88\pm{0.21}$ & $0.89\pm{0.20}$ & $0.79\pm{0.24}$ & $\emph{0.80}\pm{0.24}$\\

Base \& Thresh. & (iii) & & $0.88\pm{0.20}$ & $0.94\pm{0.18}$ & $0.76\pm{0.25}$ & $0.73\pm{0.26}$\\
Var.1 \& Thresh. & (iii) & & $\emph{0.90}\pm{0.18}$ & $\emph{0.99}\pm{0.06}$ & $0.77\pm{0.23}$ & $0.78\pm{0.21}$\\
Var.2 \& Thresh. & (iii) & & $0.88\pm{0.20}$ & $0.97\pm{0.14}$ & $\emph{0.78}\pm{0.24}$ & $\emph{0.79}\pm{0.23}$\\
Var.3 \& Thresh. & (iii) & & $0.86\pm{0.23}$ & $0.97\pm{0.15}$ & $0.77\pm{0.25}$ & $\emph{0.79}\pm{0.24}$\\
\hline
\end{tabular*}
\label{3269-tab:tprppv}
\end{table}

The detection-based results in Table \ref{3269-tab:tprppv}, which by definition also reflect FN and FP cases, supports our assumption that model Var.$1$ with the histogram matching has a better TPR compared to the other models, whereas the precision is comparable. It can be seen that the TPR for the inferior points is higher than that of the anterior points for all experiments and evaluation methods, and the PPV is comparable. Clearly, more correct inferior points are detected. This difference gets even more clear if we apply a minimal distance threshold (2nd section of Tab. \ref{3269-tab:tprppv}), here the TPR difference for Var.$1$ is $+0.09$. The threshold-based detection metrics (iii) enables a more task-dependent definition of TP cases, as we are able to include the knowledge of how accurate we need to be.
There is comparable change in the PPV but an increase in the TPR, when single points are included.

\section{Discussion}
We investigate the performance by defining multiple, goal-dependent evaluation strategies and cross-validated the performance of this pipeline which is able to handle slices without GT points. The choice of a different strategy yields a different best-performing model. Hence, associated assumptions have to be taken into account when choosing the metric. Based on our experiments, we recommend for landmark detection the usage of at least one detection and one localisation based metric. Additionally the localisation based metric needs an upper bound in the form of a FN case handling.
In this work, we have defined the upper bound on a slice-based level, meaning that the distance to the farthest corner is considered. However, this yields different values for a heterogeneous image collection of different image sizes. In future work, we will investigate these differences and compare them with a global, image-size independent, upper bound.

\begin{acknowledgement}
	The research was supported by Informatics for Life project funded by the Klaus Tschira Foundation and SDS@hd service by the MWK Baden-Württemberg and the DFG through grant INST 35/1314-1 FUGG and INST 35/1503-1 FUGG.
\end{acknowledgement}

% This command generates the bibliography using the entries of the .bib file.
% Remove it only if you do not use a bibliography. 
\printbibliography

\end{document}